\documentclass[preprint,12pt]{elsarticle}

\usepackage{amssymb}
\usepackage{multirow}
\usepackage{geometry}
\usepackage{soul}
\usepackage[normalem]{ulem}
\usepackage{amsmath}
\usepackage{graphicx}
\usepackage{booktabs}
\usepackage{multirow}
\usepackage{rotating}
\usepackage{array}
\usepackage{soul}
\usepackage{enumitem}
\usepackage{pifont}
\usepackage{picinpar}
\usepackage{color}
\usepackage{alltt}
\usepackage{float}
\usepackage[table]{xcolor}

\usepackage[colorlinks=true, linkcolor=blue, citecolor=blue, urlcolor=blue]{hyperref}
\definecolor{lightgreen}{RGB}{144,238,144}
\definecolor{myblue}{rgb}{0.8, 0.85, 1}
\definecolor{mylightgreen}{rgb}{0.9, 0.9, 0.9}
\usepackage[ruled, vlined, linesnumbered]{algorithm2e}
\sethlcolor{yellow}  

\begin{document}

\begin{frontmatter}



\title{Friction-Scaled Vibrotactile Feedback for Real-Time Slip Detection in Manipulation using Robotic Sixth Finger}

\author[inst1,inst2]{Naqash Afzal}
\author[inst1,inst2]{Basma Hasanen}
\author[inst1,inst2]{Lakmal Seneviratne}
\author[inst3]{Oussama Khatib}
\author[inst1,inst2]{Irfan Hussain\corref{cor1}}

\affiliation[inst1]{organization={Department of Mechanical and Nuclear Engineering, Khalifa University},
            addressline={Khalifa University}, 
            city={Abu Dhabi},
            postcode={127788}, 
            state={Abu Dhabi},
            country={United Arab Emirates}}
\affiliation[inst2]{organization={Khalifa University Center for Autonomous Robotic Systems (KUCARS)},
            addressline={Khalifa University}, 
            city={Abu Dhabi},
            postcode={127788}, 
            state={Abu Dhabi},
            country={United Arab Emirates}}

\affiliation[inst3]{organization={Stanford Robotics Laboratory, Computer Science Department},
            addressline={Stanford University}, 
            city={Palo Alto},
            postcode={94305}, 
            state={CA},
            country={USA}}

\cortext[cor1]{Corresponding author: Naqash Afzal. Email: naqash.afzal@bristol.ac.uk, Irfan Hussain. Email: irfan.hussain@ku.ac.ae}
\fntext[equal]{First two authors contributed equally to this work.}

\begin{abstract}
The integration of extra-robotic limbs/fingers to enhance and expand motor skills, particularly for grasping and manipulation, possesses significant challenges. The grasping performance of existing limbs/fingers is far inferior to that of human hands. Human hands can detect the onset of slip through tactile feedback originating from tactile receptors during the grasping process, enabling precise and automatic regulation of grip force. This grip force is scaled by the coefficient of friction between the contacting surface and the fingers. The frictional information is perceived by humans depending upon the slip happening between the finger and the object. This ability to perceive friction allows humans to apply just the right amount of force needed to maintain a secure grip, adjusting based on the weight of the object and the friction of the contact surface. Enhancing this capability in extra-robotic limbs or fingers used by humans is challenging. To address this challenge, this paper introduces a novel approach to communicate frictional information to users through encoded vibrotactile cues. These cues are conveyed on the onset of incipient slip thus allowing the users to perceive the friction and ultimately use this information to increase the force to avoid dropping of the object. In a 2-alternative forced-choice protocol, participants gripped and lifted a glass under three different frictional conditions, applying a normal force of 3.5 N. After reaching this force, the glass was gradually released to induce slip. During this slipping phase, vibrations scaled according to the static coefficient of friction were presented to users, reflecting the frictional conditions. The results suggested an accuracy of $94.53\pm3.05$ ($mean\pm SD$) in perceiving frictional information upon lifting objects with varying friction. The results indicate the effectiveness of using vibrotactile feedback for sensory feedback, allowing users of extra-robotic limbs or fingers to perceive frictional information. This enables them to assess surface properties and adjust grip force according to the frictional conditions, enhancing their ability to grasp and manipulate objects more effectively.
\end{abstract}







\begin{keyword}
Vibro-tactile \sep Wearable Robotics \sep Haptic Feedback \sep friction \sep extra-robotic limbs
\end{keyword}

\end{frontmatter} 

\section{Introduction}
\label{sec:sample1}
EXTRA-ROBOTIC limbs have gained significant attention over the past decade, particularly for enhancing human capabilities and, more specifically, for rehabilitation purposes. Stroke is one of the leading causes of limb paralysis, often resulting in significant and lasting impairment in the affected arm or hand  \cite{Heart_Disease_and_Stroke_Statistics—2014_Update}. This impairment can affect either the upper or lower limbs, leading to both physical and psychological strain. Consequently, it can interfere with essential activities such as walking, dressing, eating, and other daily tasks, ultimately diminishing the individual's independence and potentially causing psychological distress. Statistics from prospective cohort studies indicate that a relatively small proportion of stroke patients with upper limb paralysis achieve full recovery within six months of the event. Specifically, only between $5\%$ and $20\%$ of these individuals experience complete restoration of function in their affected arm\cite{Nakayma1994}. Thus, improving the functionality of the affected hand is essential for the overall recovery of stroke patients with upper limb paralysis. This enhancement not only contributes significantly to their ability to perform daily activities but also plays a vital role in their overall rehabilitation process\cite{FARIAFORTINI2011257, Kwakkel2007}. As a result of this trauma, various motor impairments can significantly impact hand function, affecting both motor execution and motor planning/learning. These impairments may include weakness in the wrist and finger extensors, increased muscle tone and spasticity in the wrist and finger flexors, cocontraction of muscles, where opposing muscles contract simultaneously, impairing smooth movement, reduced independence of finger movements, poor coordination between grip and load forces, making it difficult to manage objects effectively, inefficient scaling of grip force and peak aperture, leading to difficulties in grasping and manipulating objects, delays in the preparation, initiation, and termination of gripping actions, affecting the ability to handle objects safely \cite{00019052-201012000-00019}. These motor impairments highlight the complex nature of hand function recovery after stroke and underscore the need for targeted rehabilitation strategies to address each specific issue. 

Humans have developed and harnessed tools to amplify their sensory and motor abilities. This remarkable aptitude for tool use is a signature trait of our species, profoundly shaping our evolution and advancement throughout history. Extending sensory processing beyond the nervous system is common in the animal kingdom; for instance, rodents use whiskers to explore objects, while spiders detect prey through their webs \cite{andy1998extended}. Similarly, in humans, the nervous system perceives tools as extensions of the body's sensory system, rather than mere connections between the hand and the environment \cite{HOLMES200462, miller2018sensing, Millerarticle}. Building on this natural phenomenon, modern technology has advanced robot-assisted therapies, which are now widely adopted in rehabilitation and have gained significant attention for their seamless integration into treatment plans \cite{Volpe, 5771114}. The use of extra-robotic limbs or fingers has been particularly effective, as evidence shows that their application leads to enhanced neural representations and improvements over time \cite{doi:10.1126/scirobotics.abd7935}. Robotic devices enable high-intensity, repetitive, and task-specific exercises, delivering interactive treatment to the affected limb. Additionally, these technologies offer an objective and reliable means of tracking patient progress throughout the rehabilitation process. Furthermore, motor augmentation, a rapidly advancing field, is dedicated to enhancing human physical capabilities beyond rehabilitation. Consequently, researchers are currently developing advanced supplementary robotic fingers and limbs designed to provide functional support. These augmentative devices are designed to transform our interaction with the environment, leading to alterations in how we move and control our biological bodies. By integrating with our physical capabilities, they enhance and redefine our physical interactions and operational efficiency\cite{7406744, Wu, Shafti2021}. Recently, Dominijannia \textit{et al.} addressed a critical and previously unresolved challenge: allowing users to control extra-robotic limbs effectively without disrupting their existing functional abilities\cite{scirobotics.adh1438}. Users can receive training to master the use of extra-robotic limbs while seamlessly integrating them with their routine motor tasks. This proficiency enables them to perform complex activities with greater efficiency and effectiveness. Despite the rapid progress in the development of extra-robotic limbs, a critical aspect often overlooked is how to effectively convey sensory feedback from the limb-object interface to the user, particularly during gripping and manipulation tasks. 

In order to match the sensory capabilities of human fingers, providing sensory feedback from the fingers to the user is essential, especially when robotic limbs are used for gripping and manipulation tasks. When we hold objects like tools, in addition to automatic motor adjustments, we consciously sense the slipperiness of the surface and whether the skin provides enough traction for safe and effective handling. Therefore, sensing friction between the object and the finger is crucial for maintaining safe and secure contact during manipulation. Grip force regulation using extra-robotics limbs is quite challenging because of the dynamic nature of the task. In the present study, we developed a novel technique to detect real-time slip during object lifting with extra-robotic fingers. This approach utilizes vibrotactile cues to deliver real-time feedback to the same limb, with the intensity of the cues dynamically scaled to the coefficient of friction between the robotic sixth finger and the object as soon as slip is detected (Fig. 1). We further evaluated the strategy using a two-alternative force choice protocol by using three different levels of friction to the participants. Our results indicated $94.53\%$ overall accuracy in perceiving the three different levels of friction using the robotic sixth finger.   

\begin{figure}[!t]       
    \includegraphics[width=8.5cm]{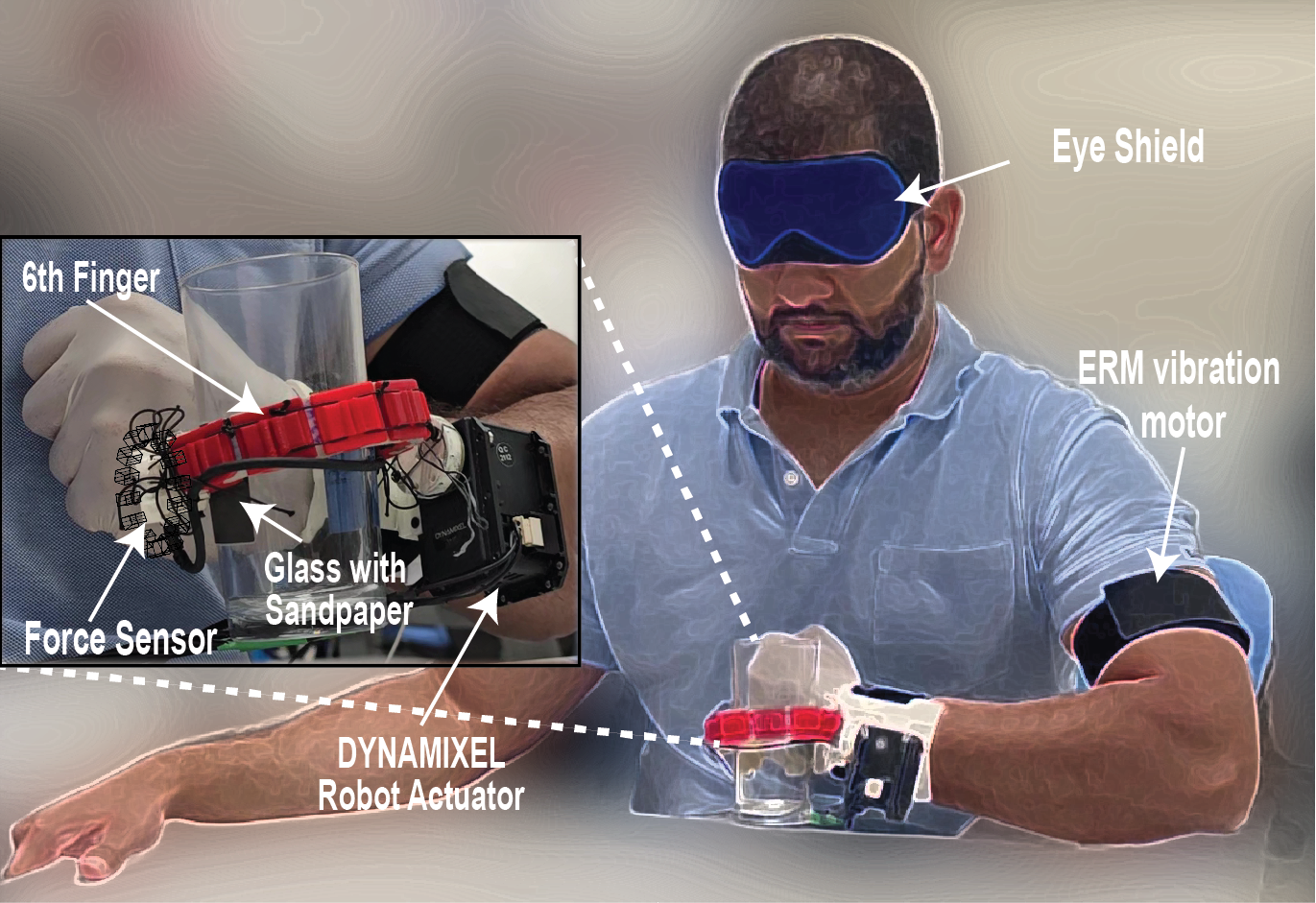}
    \centering
    \caption{Instrumented Robotic Sixth Finger: Participant performing griping and lifting task using the instrumented robotic sixth finger}
    \label{FIG_1}
\end{figure}

\section{Related Work}
Supernumerary robotic limbs (SRLs) are an emerging category of wearable auxiliary devices designed to assist humans, and they have recently become a major area of interest in research around the world. Various types of SRLs have been developed to serve different functions, acting as an extra arm or leg to support the human body and assist with tasks or operations. These applications include using robotic arms to assist aircraft assembly workers in handling heavy tools \cite{6906601, 7410066}, providing additional fingers to help hemiplegia patients complete everyday tasks \cite{HUSSAIN20171}, and employing robotic arms to work in coordination with users to perform complex or fatigue-inducing tasks, such as overhead operations \cite{6386055}. SRLs provide essential physical support to their users, offering new possibilities to enhance human motor skills. However, grasping is one of the most fundamental and critical functions of extra-robotic upper limbs and fingers which remains a significant challenge. Reliable force control during object grasping is essential. Applying too much force can damage the object, while insufficient force might cause the object to slip. Effective grasping requires adjusting grip strength based on the friction between the fingers and the object. Although many grippers are pre-programmed to handle specific tasks, this approach struggles in unstructured environments where the characteristics of objects are unknown in advance, making precise grasping more challenging. An effective grasping control strategy should allow a robot to handle objects with varying characteristics within its operational limits. Relying on maximum force for grasping is impractical, as it can damage delicate objects. In this regard, current robotic grippers' grasping capabilities fall short compared to human hands. Humans can accurately adjust grip force based on the tactile feedback from their fingers, ensuring that the force applied does not significantly exceed what is required for a safe grip \cite{westling1984factors}. These capabilities significantly enhance the range of objects that human hands can grip and manipulate. Previous research has shown that detecting slips at the fingertip is crucial for controlling grasping force effectively \cite{johansson2009coding}. 
The human hand regulates grip force by detecting friction between the fingers and the contact surface through subtle fingertip movements or slip displacement \cite{johansson1984roles, johansson1987signals, https://doi.org/10.1113/JP286027}. This slip sensation in human hands is detected before larger, more noticeable slips occur \cite{9669043}. Hence, integrating slip-sensing capabilities into robotic grippers is crucial for improving their grasping performance, as it mirrors the human ability to sense and adjust grip before significant slipping occurs. 

The slip hypothesis in epicritic tactile perception suggests that when sensors and objects move relative to each other, they act like a frictional system, causing sudden, jerky movements called 'slips'. These slips are influenced by the object's shape, the forces involved, the materials in contact, and the surrounding environment. This process helps encode sensory information and shapes how we use our perceptual strategies, including how we move our sensors \cite{SCHWARZ2016449}. In human hands, all tactile afferents are responsive to slip events, but fast-adapting type I (FA-I) afferents, in particular, accurately encode the compressive strain rates associated with these slips. Due to their high density in the fingerpads, FA-I afferents are especially effective at detecting incipient slips \cite{delhaye2021high}. Slip detection involves two main phases: macro slip, where the entire surface of two objects slides past each other, and incipient slip, where only some parts of the surface move while others stay in place. Numerous reviews provide comprehensive overviews of the various methods used for slip detection \cite{francomano2013artificial}, \cite{chen2018tactile}. 
These reviews reveal that macro slip detection is limited to indicating just two states: slip or no-slip. When macro slip is absent, the data fail to reveal how close the situation is to macro slip, restricting its effectiveness in guiding grip force control. In contrast, human hands can adapt their grip by reducing excessive force, especially when handling very light objects, to optimize effort and avoid unnecessary strain \cite{Westling, 7139709}. Moreover, relying on macro slip detection alone, which only starts adjusting grip force after a slip has occurred, makes it challenging to apply sufficient force to prevent the object from falling. In contrast, human hands can sense friction conditions on the object's surface and adjust grip force proactively, preventing macro slip before it happens \cite{doi:10.1073/pnas.2109109118,9669043}. 
In summary, relying solely on the macro slip for controlling grip force lacks both precision and reliability.

Most current research on detecting incipient slip qualitatively has focused on vibration analysis \cite{99981, 5976472, 509205}. However, these studies fail to distinguish between the vibrations associated with incipient slip and those occurring during macro slip.  Additionally, some studies for qualitative detection of incipient slip utilize strain measurements \cite{976249}, analyze contact regions \cite{s18020333}, or rely on specialized sensor designs \cite{doi:10.1080/15599610701232655}. However, few studies offer direct validation of whether their detection methods truly identify incipient slips. While many claim that their techniques can detect incipient slip before macro slip \cite{99981,5976472,509205,976249} they often do not specify or verify the exact timing of this detection. Hence, Chen \textit{et al.} have raised concerns about whether these methods truly detect incipient slip or simply identify it earlier than conventional macro slip detection techniques. Even if these methods can capture the incipient slip state before macro slip occurs, they only offer an improvement over macro slip detection. They still fall short in providing information about how close the contact state is to macro slip when incipient slip is not detected. Therefore, they remain inadequate for tasks requiring precise grip force control \cite{chen2018tactile}. Quantitative detection of incipient slip provides a precise measurement of how close the current contact state is to macro slip by using specific indicators. By continuously monitoring this data, we can fine-tune the grip force to keep the degree of incipient slip near a critical threshold, just before macro slip occurs. This approach enables the application of the optimal amount of force, avoiding excessive pressure while maximizing the success rate of grasping tasks. Such control closely mirrors the precision of manual grip adjustments, which are based on real-time tactile feedback. Therefore, in this paper we aim to enhance the detection of incipient slip and integrate haptic feedback mechanisms, ensuring that the robotic sixth finger is equipped with sensory capabilities rather than functioning solely as a mechanical assistive device. By integrating these advancements, we aim to enhance the adaptability and effectiveness of supernumerary robotic fingers, making them more responsive and intuitive in handling a diverse range of objects.

Vision-based techniques are widely utilized to assess the extent of incipient slip. The current methods for assessing incipient slip focus on detecting stick and slip zones on the contact surface, using the ratio of the stick (or slip) region to the total contact area as a key metric for gauging the extent of slip. Maeno \textit{et al.} measured the shear strain distribution within silicone during contact by embedding strain gauges into the material \cite{976249}. They showed that this distribution could accurately represent the stick-slip condition on the contact surface. Additionally, the stick and slip regions were identified by analyzing the strain changes at each strain gauge \cite{845338}, and the grip force was subsequently adjusted based on the size of the stick region. Vision-based tactile sensors are gaining popularity for slip detection. These sensors are integrated with an internal camera and feature multiple embedded markers on their surface. By tracking the movements of these markers, the sensors can capture detailed deformation data from the contact surface \cite{4420900, 8202149, 8793538, SUI2022111906, 9565930}. This approach, which relies on the area ratio of the stick and slip regions, has a well-defined physical interpretation. It provides a direct measure of how close the system is to experiencing macro slip, as the complete disappearance of the stick region signals the onset of macro slip. Furthermore, this approach is effective in cases involving rotational and translational slips \cite{SUI2022111906,9812186}. This is because the identification of the stick/slip region is typically unaffected by the specific type of slip. However, this method is limited to simple contact scenarios and becomes ineffective when the direction of the tangential force changes or macro slip occurs, making it difficult to accurately assess the degree of incipient slip. Additionally, it is computationally intensive due to the complex image processing required, which poses a significant challenge. In summary, while quantitative slip detection methods have been implemented in robotic grippers, integrating extra-robotic limbs with slip detection for automatic grip force regulation presents significant challenges that have not been adequately addressed. To tackle this, our paper proposes a novel technique for detecting slip and encoding this information in the form of vibrotactile cues. This sensory feedback is then communicated to the user to sense friction of the object, which will further enable extra-robotic limbs to adjust grip force dynamically and optimize stability when handling lightweight objects in the future.

\begin{figure}[!t]
\centerline{\includegraphics[width=\columnwidth]{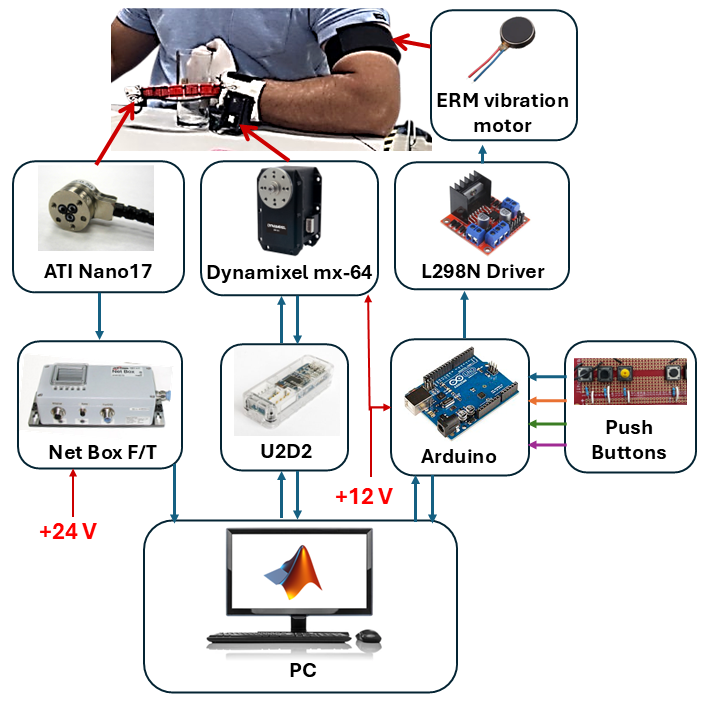}}
\caption{Schematic of the whole control process after getting force feedback from the sensor and sensing haptic feedback to the user}
\end{figure}

\section{Methods}
\subsection{Ethical Approval}

The experimental protocols were approved by the Human Research Ethics Committee (Approval \# H23-007) at Khalifa University, Abu Dhabi. The study conformed to the standards set by the Declaration of Helsinki, except for registration in a database. All the subjects provided written consent before the start of the experiment.

\subsection{Participants}

Eight healthy right-hand-dominant subjects (7 male (age $33.87$ ± $6.35$ years, $mean$ ± $SD$); 1 female) participated in the experiment. All subjects in the study reported no history of neurological disorders and presented no clinical signs that would indicate altered skin sensitivity or motor function of the hand. All subjects used their non-dominant hand for the experimental tasks.

\subsection{Experimental apparatus}

A 3D-printed robotic sixth finger, as developed by \cite{7406744}, features a modular flexible finger, a support base, and an adjustable strap for attachment to the user’s arm. This design utilizes a combination of 3D-printed PLA polymer for structural rigidity and thermoplastic polyurethane for flexibility. The finger's movement is driven by a single actuator, which pulls a tendon (or fishing wire) running through the rigid section of the finger. The tendon is anchored at one end to the fingertip and at the other to a pulley on the actuator, ensuring a stable grip. The finger automatically adapts to variations during use, improving grip efficiency. The support base houses the actuator and includes a strap that secures the device to the user’s forearm, with a symmetrical design allowing for use on either arm. A custom 3D-printed housing was attached to the fingertip to mount a six-axis ATI Nano17 force-torque sensor (ATI Industrial Automation, NC, USA). The finger's actuation is powered by a Dynamixel MX-64T servo motor (Robotis Co. LTD, Seoul, Korea), controlled via an ArbotiX-M Robocontroller. For haptic feedback, an armband was equipped with an ERM (Eccentric Rotating Mass) coin vibration motor (Precision Microdrives Co. LTD, USA), which is operated using an L298N Motor Driver Module controlled by an Arduino UNO (Fig. 2).  An external battery is used to provide power to all the circuits. All the electronics are enclosed in a 3-D printed box. A transparent water glass was mounted with two different sandpapers with grit no. P5000 represents the medium friction surface in the experiments and P800 represents the High friction surface in the experiments. The weight of the glass was about 280g. A soft PLA-printed base was affixed to the "Tool Adapter Plate" of the force sensor, which made contact with the gripping object. To measure friction, we generated slip by gradually releasing the glass, and assessed the friction between the soft base on the sensor's fingertip and the glass surfaces. The static coefficient of friction ($\mu_s$) was determined by calculating the ratio of tangential to normal force at the point when the entire fingertip contact area began to slip. Slip timing was identified through visual inspection of the force traces, specifically when a decrease in tangential force was observed (Table 1).

\begin{table}[h]
\caption{Measured coefficient of friction}
\label{table}
\setlength{\tabcolsep}{3pt}
\centering
\begin{tabular}{|c|c|c|}
\hline
\textbf{Friction level} & \textbf{Contacting surface} & 
\textbf{COF ($\mu$)} \\
\hline
Low & Glass & 
$0.25$ \\
\hline
Medium & Sandpaper (grit no. P5000) & 
$0.55$ \\
\hline
High & Sandpaper (grit no. P800) & 
$0.95$ \\
\hline
\end{tabular}
\end{table}

\subsection{Experimental protocol}

Participants were seated comfortably in a height-adjustable chair, with the robotic sixth finger attached to the distal forearm on the volar side, near the wrist. The device was supported by the volar part of the wrist, aiding in lifting objects. An armband equipped with an ERM vibration motor was worn on the biceps of the same arm to provide haptic feedback. To eliminate visual cues, participants wore an eye shield, preventing them from seeing the glass gripped by the sixth finger. The task required participants to grip and lift a glass in different orientations, each with varying friction levels at the fingertip (Fig. 3A). A two-alternative forced-choice (2AFC) protocol was used in the psychophysics study to evaluate participants' ability to perceive slip via haptic feedback on the biceps. For each pair of trials, two grip-and-lift actions were performed with different friction levels. The experimenter adjusted the glass orientation and initiated the gripping task using a graphical user interface (GUI) developed in MATLAB, which displayed commands and force targets via real-time progress bars (Fig. 3B). 

The trial began with the command, "Press Button-1 to initiate gripping." The experimenter confirmed the correct orientation of the glass’s contact surface with the fingertip before pressing Button-1 to start the finger’s motion. The finger continued to move until it touched the surface of the glass, where a short vibration cue was delivered to the participant upon initial contact, detected when the grip force reached 0.5N. Next, the GUI prompted, "Press Button-2 to continue gripping until 3.5N." The experimenter pressed Button-2, increasing the grip force to the target threshold of 3.5N. Once reached, the GUI instructed, "Press Button-3 to slowly release the glass." The participant was directed to lift the glass 2-3 cm above the table and hold it in the air. The experimenter then pressed Button-3 to initiate a slow release. When the glass began to slip, a vibrotactile cue, scaled according to the friction between the glass and the fingertip, was delivered to the participant’s forearm. After the vibration stopped, the experimenter pressed Button-4 to fully release the object. The next trial, with a different friction level and glass orientation, followed the same procedure. Once both lifts in the pair were completed, participants verbally indicated which of the two stimuli felt more slippery (i.e. which vibration cue had less intensity) (Fig. 3D), and their responses were recorded by the experimenter.

\begin{figure*}[!t]
\centering
\includegraphics[width=\linewidth]{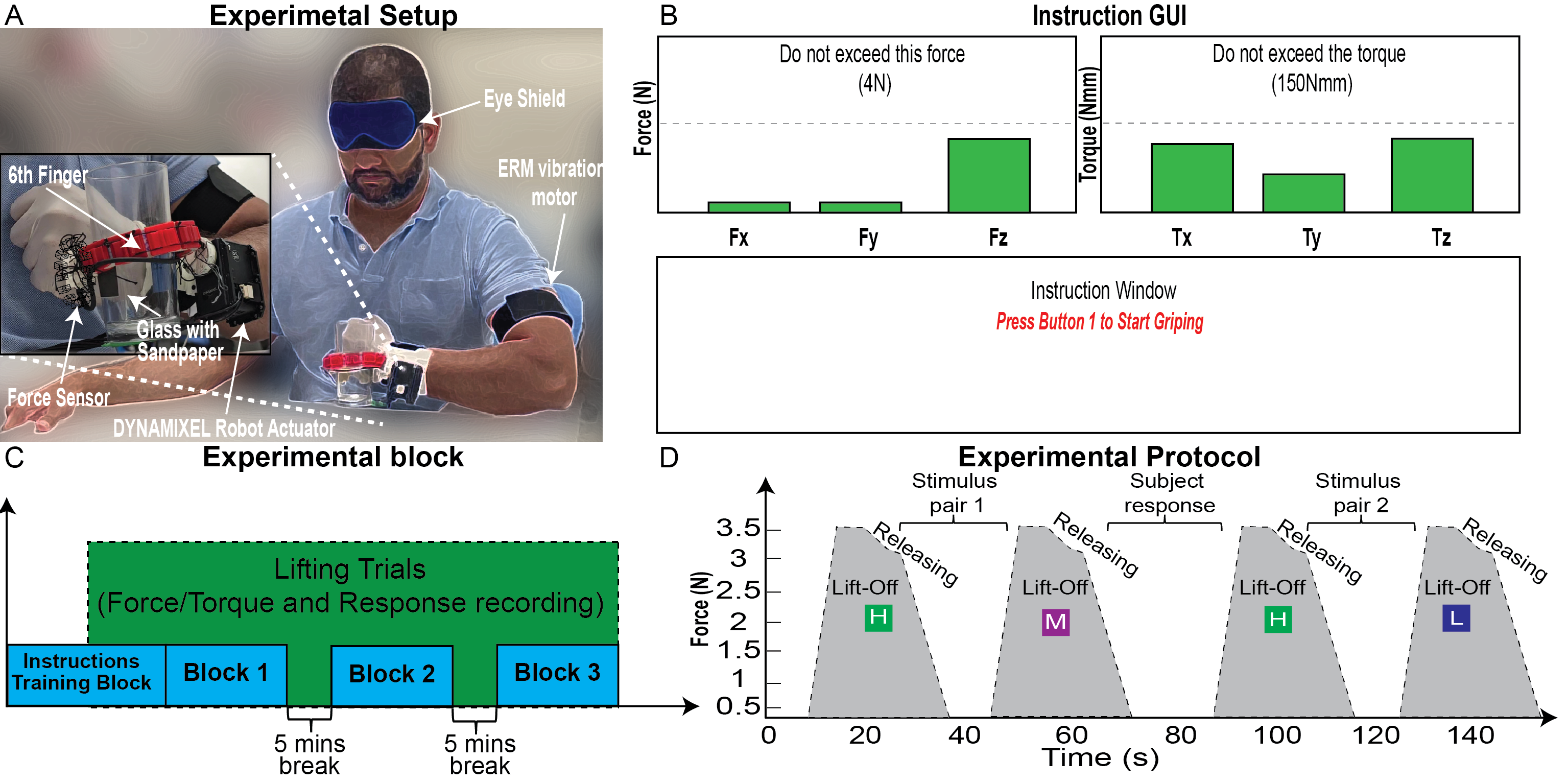}
\caption{Experimental setup and protocol: A, Participant wearing robotic sixth-finger instrumented with an ATI-Nano17 Force-torque sensor, an armband with an ERM vibration motor, and an eye shield B, GUI designed to guide the experimenter during trials where the sixth finger grips a glass with three different friction levels. The GUI prompts the experimenter to initiate the grip and, after reaching a threshold normal force of 3.5N, lift the glass. Release begins after lift-off. Real-time bar charts display force/torque levels to ensure target forces are met and to prevent damage during process C, the schematic of the experimental sequence D, schematic illustration of the time course of presentation and evaluation of the stimulus pairs by subjects touching the friction modulation device (H vs. M and H vs. L denote pairs of stimuli where H is high friction; L is low friction; M is medium friction). Glass is gripped for the L condition and sandpaper with different grit no. is gripped for the M and H condition.}
\end{figure*}

A training block consisting of 10 stimulus pairs was conducted before the experiment to familiarize participants with the task. The 2-AFC protocol was used to assess participants' ability to perceive frictional differences through haptic feedback delivered to the forearm. Each condition was tested using 30 stimulus pairs, divided into three experimental blocks of 10 pairs each, with a 5-minute break between blocks to prevent bicep numbness and stimuli adaptation. Each pair of stimuli (H–M: high vs. medium, M–L: medium vs. low, H–L: high vs. low) was presented 10 times. In 5 trials, the higher-friction stimulus was presented first, followed by the lower-friction stimulus, and in the remaining 5 trials, the lower-friction stimulus came first. All stimuli were presented in a pseudorandom order. A schematic of the entire experimental protocol is shown in Fig. 3C.

\subsection{Statistical analysis}

Analysis of variance (ANOVAs) and post-hoc paired sample tests (with Bonferroni corrected values for multiple comparisons) were performed if a comparison of more than two groups was required. Fractional degrees of freedom are reported accordingly to the Greenhouse–Geisser correction when Mauchly's test of sphericity showed that the assumption of sphericity had been violated. When a D'Agostino–Pearson normality test \(P < 0.05\) indicated that the data was not normally distributed, instead of ANOVA, the Friedman test was used. 

\subsection{Data Acquisition}
Signals from the force/torque sensor were captured using MATLAB (R2023b) and the Net/FT Box integrated with a desktop computer equipped with an Intel i9-14900KF processor. The position data from the Dynamixel MX-64T servo motor, which actuates the tendon-driven mechanism controlled by the ArbotiX controller, were also recorded using MATLAB and saved in CSV format. All force and torque measurements, as well as motor position data were systematically recorded and stored in CSV files for subsequent analysis.

\subsection{Algorithm for friction-scaled haptic feedback on the onset of slip}
In the gripping process, two primary forces are crucial for stabilizing the grip and ensuring successful manipulation. The grip force, or normal force, is applied perpendicular to the object being gripped and is denoted by $f_n$. During the lifting of the object, a tangential force $f_t$ is generated at the interface between the fingertip and the contacting surface. This tangential force is scaled by the coefficient of friction between the surface and the weight of the object and acts in the direction opposite to the object's weight.

The ratio of the magnitude of the tangential force to the normal force is denoted as the $\mu_s$:

\[
\mu_s = \frac{|f_t|}{|f_n|}
\]

Considering only Coulomb friction, the minimum tangential force required to maintain a secure grip on the object can be determined. This relationship is defined as:
\[
f_t \geq \mu_s f_n
\]
This inequality ensures that the tangential force is sufficient to counteract the frictional resistance and prevent the object from slipping. However, it will be challenging if the $\mu_s$ is not known because $\mu_s$ can be measured only at the moment of slip onset, thus we devised instead to use the acceleration of tangential force to devise the slippage. To detect slip, the double derivative (acceleration) of the tangential force \( \frac{d^2f_t}{dt^2}\) is monitored. A quantitative measure for \( \frac{d^2f_t}{dt^2}\) greater than the threshold value is set to indicate slip. After conducting pilot experimentation we devised the threshold to be \( 0.3 \,{N}/{s^2} \).


If the second derivative of the tangential force exceeds a threshold, it is interpreted as an indication of slippage. Consequently, upon detecting this event, the frequency/amplitude of the vibration motor delivering haptic feedback is scaled according to the relationship (\ref{eq:combined_frequency}):

\scriptsize
\begin{align}
    \max &\left(30, \min\left(255, \frac{f_{\text{peak}} \cdot \frac{1}{SR_{\text{peak}}}}{-\Delta d^2f_t}\right)\right), \text{if } \Delta d^2f_t < 0 \ (\textit{Low}) \label{eq:frequency_low} \\
    \max &\left(30, \min\left(255, f_{\text{peak}} \cdot \frac{1}{SR_{\text{peak}}} \cdot \Delta d^2f_t\right)\right), \text{if } \Delta d^2f_t > 0 \ (\textit{High})
    \label{eq:frequency_high}
\end{align}
\small

The final frequency \(f\) is the minimum value derived from both conditions:

\begin{equation}
A/f = \min \left(
\begin{aligned}
&\eqref{eq:frequency_low} \\
&\eqref{eq:frequency_high}
\end{aligned}
\right)
\label{eq:combined_frequency}
\end{equation}

where:
\begin{itemize}[itemsep=3pt]
    \item \(f_{\text{peak}}\) is the peak frequency of the vibration motor (255).
    \item \(SR_{\text{peak}}\) is the peak slip ratio observed after a stable grip.
    \item \(\Delta d^2f_t\) is the second derivative difference of the tangential force.
\end{itemize}

Simultaneously, the change in slip ratio is continuously monitored. When the slip ratio difference exceeds a threshold, it serves as another indicator of slippage. As a result, the frequency and amplitude of the vibration motor providing haptic feedback are scaled according to the relationship (\ref{eq:slip_ratio}): 

\begin{equation}
A/f = 
\begin{cases} 
\min\left(30, \max\left(255, \frac{f_{\text{peak}} \cdot k}{SR_{\text{peak}}}\right)\right) & \text{if } \Delta SR \geq 0.5 \\
0  & \text{if } \Delta SR < 0.5
\end{cases}
\label{eq:slip_ratio}
\end{equation}

where:
\begin{itemize}[itemsep=3pt]
    \item \(k\) is a constant that adjusts the scaling factor to ensure the frequency remains within bounds.
\end{itemize}

\begin{algorithm}[htbp]
\scriptsize
\DontPrintSemicolon
\KwIn{$f_n$: Normal Force, $f_t$: Tangential Force}
\KwOut{$A,~f$}
\KwData{%
        $K$: Control Gain, $A_{\text{max}}$: Maximum Amplitude, $f_{\text{max}}$: Maximum Frequency
    }{}
$\left( \frac{d^2f_t}{dt^2} \right)_{\text{thresh}} = 0.3~N/s^2$: Tangential Force Acceleration Threshold \;
\colorbox{red!10}{Initialization:} \;
Set initial amplitude $A = 0$ and frequency $f = 0$ \;
Set timer $t_{\text{start}} = 0$ to start 10 seconds after peak tangential force \;
\colorbox{red!10}{Detect Peak Normal Force:} \;
\If{$f_n$ reaches its peak value}{
    Set $t_{\text{start}} = \text{current time} + 10$ seconds
}
\colorbox{red!10}{Begin monitoring after 10 sec:} \;
\While{$\text{Sensor data} > 0$ and $\text{current time} > t_{\text{start}}$}{
    Compute $\frac{df_t}{dt}$ and $\frac{d^2f_t}{dt^2}$ from sensor data \;
    Compute current slip ratio $SR = \frac{f_n}{f_t}$ \;
    \colorbox{gray!20}{Check Double Derivative:} \;
    Compute the difference in double derivative:
    $\Delta d^2f_t = \left( \frac{d^2f_t}{dt^2} \right)_{\text{thresh}} - \frac{d^2f_t}{dt^2}$ \;
    \uIf{$\Delta d^2f_t > 0$}{
    $A/f \gets \min(A/f_{\text{max}}, K \cdot \Delta d^2f_t)$ \;    
    } 
    \Else{
    $A/f \gets \max \left(0, K \cdot (-\Delta d^2f_t)\right)$ \;
    
    }  
    $A/f \gets \min \left(
    \begin{aligned}
        &\max\left(30, \min\left(255, \frac{f_{\text{peak}} \cdot \frac{1}{SR_{\text{peak}}}}{-\Delta d^2f_t}\right)\right), \\
        &\max\left(30, \min\left(255, f_{\text{peak}} \cdot \frac{1}{SR_{\text{peak}}} \cdot (-\Delta d^2f_t)\right)\right)
    \end{aligned}
    \right)$
    \colorbox{gray!20}{Check Slip Ratio:} \;
     Compute the difference in slip ratio: $\Delta SR = |SR_{\text{peak}} - SR|$
 
    \uIf{$\Delta SR \geq 0.5$}{  
    $A/f \gets \min\left(30, \max\left(255, \frac{f_{\text{peak}} \cdot k}{SR_{\text{peak}}}\right)\right)$
    }
    \Else{  
    $A/f \gets 0$
    }
    \colorbox{gray!20}{Simultaneous Condition Check:} \;
    \uIf{$\left( \Delta d^2f_t > 0 \right) \ \textbf{or} \ \Delta SR > 0.5$}{
        Adjust amplitude and frequency based on the condition met first \;
        \uIf{$\Delta d^2f_t > 0$}{
            Adjust amplitude and frequency based on double derivative
        }\Else{
            Adjust the frequency based on slip ratio difference
        }
        Send the signal to the vibration motor with the values $A$ and $f$
    } \Else{
        \colorbox{green!20}{Motor Silent:}
        $A \gets 0$, $f \gets 0$ \;
        No signal sent to vibration motor
    }
    \colorbox{gray!20}{Update Parameters:} \;
    Recalculate $\frac{d^2f_t}{dt^2}$ and $SR$ using the latest sensor data \;
    \If{$A = A_{\text{max}}$ or $f = f_{\text{max}}$}{
        Reduce $K$ to prevent oscillation
    }
    \colorbox{gray!20}{Log Data:} \;
    Store current values of $A$, $f$, $\frac{d^2f_t}{dt^2}$, and $SR$ for analysis
}
Final values of amplitude $A$, frequency $f$, and corresponding force rate $\frac{d^2f_t}{dt^2}$ \;
 
\caption{Algorithm for friction-scaled haptic feedback.} \label{alg:1}
\end{algorithm}

\section{Results}

First, we computed the static coefficient of friction between the fingertip of the 6th finger and the contacting surface. We ensured there was a significant difference in the friction coefficient between the pair of lifting trials. The coefficient of friction for the three surfaces glass, and 2 sandpapers are given in Table I.

\subsection{Ability to perceive slip}

The ability to perceive friction of the individual subjects for the three pairs of stimuli is shown in Fig. 4. The overall accuracy across subjects was $94.53 \pm 3.05\% (mean \pm SD; n = 8)$. For the pairs of stimuli with the largest frictional difference (H–L), all 8 subjects performed $100\%$. The discrimination performance between the two stimuli with a smaller frictional difference, H–M and M–L, also remained very high and was $83.75 \pm 9.16\% $ and $98.75 \pm 3.54\% (mean \pm SD; n = 8)$, respectively. There was a significant difference between the performance of the subjects among the three pairs of stimuli (one-way repeated-measures ANOVA, $(F(2,21) = 20.351, P < 0.5)$. Post hoc analyses indicated that the performance differed between stimulus pairs of highest frictional difference ($H\text{-}M vs. M\text{-}L, P < 0.5, H\text{-}M vs. H\text{-}L, P < 0.5$) and was not significant between pairs of intermediate differences ($H\text{-}M vs. M\text{-}L, P = 0.62$; Bonferroni-corrected multiple comparisons). Only one subject, S2 has a performance of 90\% for the M-L stimulus pair. Even though only one subject S4 performed 100\% during the H-M pair of stimuli all other subjects were also above 80\%. 

The experimental results are presented in a 3x3 stimulus-response confusion matrix (Table II). The rows represent the actual intensity of the vibrotactile stimulus, while the columns indicate the perceived intensity reported by the participants. Each cell in the matrix shows the frequency with which a given stimulus was classified as a specific intensity.  Table II lists the data pooled from all the 8 participants. Accuracy for the intensity perception ranged $85.6\% (High Friction)$, $99.37\% (Medium Friction)$ and $100\% (Low Friction)$. A visual inspection of the confusion matrix depicts that the maximum number of trials falls in differentiating between the High friction and the medium friction. 

\begin{figure}[!t]
\centerline{\includegraphics[width=\columnwidth]{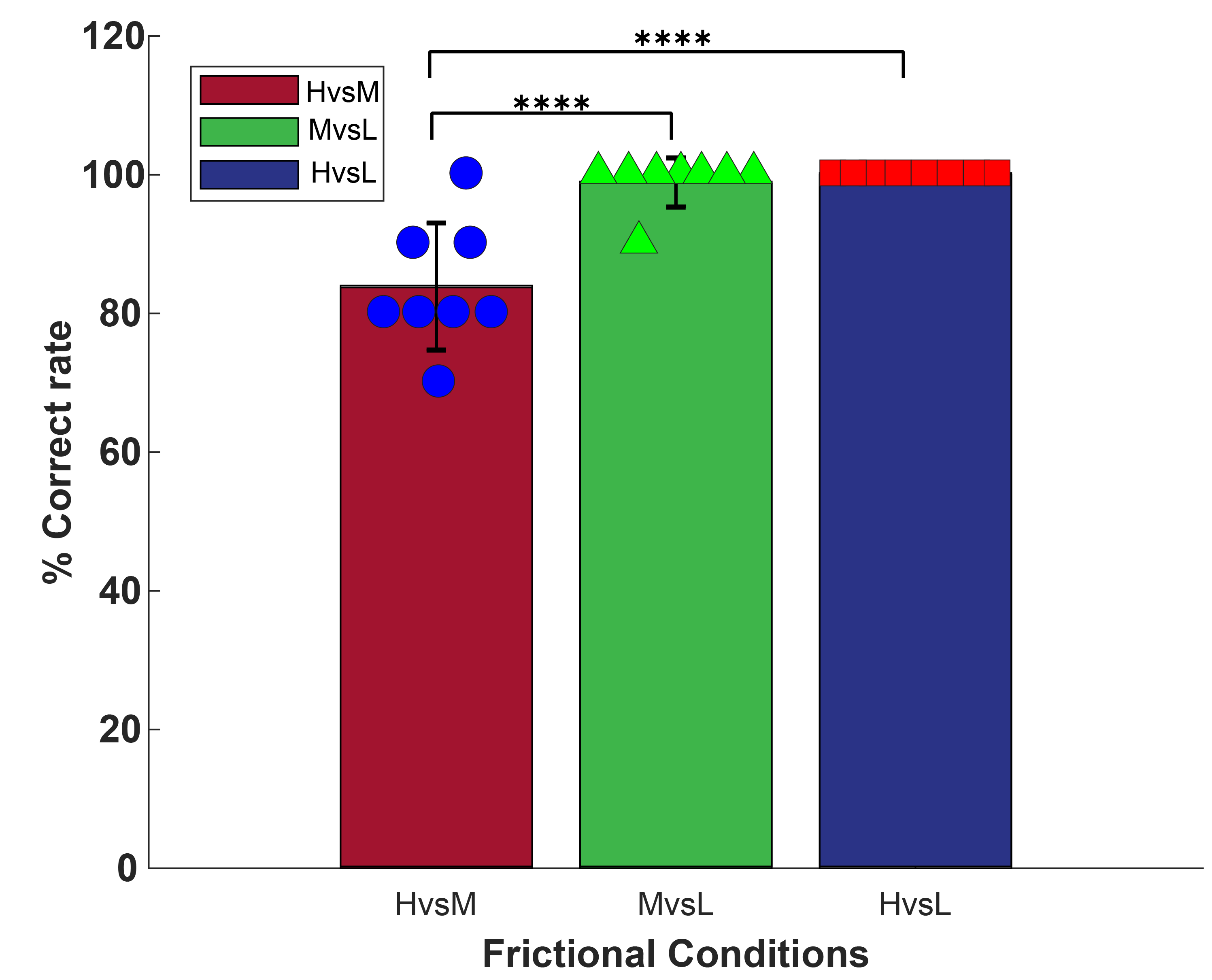}}
\caption{Participants performance: Barplots displaying the mean and standard deviation percentages of correct responses across friction pairs. $****P < = 0.00001$ (Bonferroni corrected). Blue circles show the individual responses for the HvsM condition, green triangles show individual responses for the MvsL condition and red squares show individual responses for the HvsL condition}
\end{figure}

\begin{table}[!b]
    \renewcommand{\arraystretch}{1.3}
    \caption{Confusion Matrix for the vibrotactile magnitude perception}
    \centering
    \label{table_3}
    \tiny
	\resizebox{\columnwidth}{!}{
	\begin{tabular}{ccccc}        
        \toprule        
        \multicolumn{5}{c}{\textbf{Response}} \\
        \toprule
        \textbf{Stimulus} & \textbf{High} & \textbf{Medium} & \textbf{Low} & \textbf{Sum} \\
        \midrule
        \cellcolor{mylightgreen}\textbf{High} & \cellcolor{myblue}\textbf{137} & \cellcolor{mylightgreen}23 & \cellcolor{mylightgreen}0 &  \cellcolor{mylightgreen}\textbf{160} \\
        \cellcolor{mylightgreen}\textbf{Medium} & \cellcolor{mylightgreen}0 & \cellcolor{myblue}\textbf{159} & \cellcolor{mylightgreen}1 &  \cellcolor{mylightgreen}\textbf{160} \\
        \cellcolor{mylightgreen}\textbf{Low} & \cellcolor{mylightgreen}0 & \cellcolor{mylightgreen}0 & \cellcolor{myblue}\textbf{160} & \cellcolor{mylightgreen}\textbf{160} 
        \\
        \hline
        \cellcolor{mylightgreen}\textbf{Sum} & \cellcolor{mylightgreen}\textbf{137} & \cellcolor{mylightgreen}\textbf{182} & \cellcolor{mylightgreen}\textbf{161} &  \cellcolor{myblue}\textbf{480} \\
        \bottomrule
    \end{tabular}
    }
\end{table}

\subsection{Friction-Scaled tangential forces during grasping}

The normal force exerted by the robotic sixth finger is applied perpendicularly to the object, ensuring a secure grip, while the tangential force increases simultaneously (Fig. 5). The target normal force applied to the glass was consistently maintained at $3.5N$ across all conditions before lift-off. This specific value was determined based on the weight of the glass, combined with repeated pilot tests conducted on different surfaces. The pilot tests ensured that the chosen normal force was optimal, allowing for successful and reliable lifting on the three selected surface types, while preventing unnecessary slippage or failure during the lifting process. Additionally, maintaining a constant force helped standardize the experiment, ensuring that the variations in performance were due to other experimental factors, such as surface characteristics, rather than inconsistencies in the applied force.

The inset in each graph (Fig. 5) represents the mean and standard deviation of the tangential forces, which are synchronized at a baseline of $0.5N$. These tangential forces developed between the fingertip of the robotic sixth finger and the contacting surface, are scaled according to the coefficient of friction of the surface in contact. An analysis of individual trials was conducted on three different surfaces—Sandpaper 1 (high friction), Sandpaper 2 (medium friction), and Glass (low friction) and demonstrated that the material in contact with the fingertip had a significant effect on the tangential force generated, even though the weight of the object remained constant throughout the trials. The results showed that as the surface became more slippery, the tangential force required to maintain a secure grip and initiate lifting was notably reduced. This effect was particularly evident during the gripping and lifting phases, where the nature of the contact surface played a critical role in influencing the force dynamics. It appeared that forces deviated earlier in the case of low friction surface contact as compared to the high and medium friction surface contact. 

In addition to assessing the overall tangential forces, we conducted a detailed analysis of the peak tangential forces that were developed during the lifting trials (Fig. 6 (Bar graphs peak Tangential force)). The peak tangential force provides crucial insights into the interaction between the robotic sixth finger and the varying surface materials, as it reflects the maximum force developed to counter gravity and avoid dropping of the object during the critical phase of object lifting. The average peak tangential forces measured across all subjects were $1.47 \pm 0.18N$ for the high-friction surface (Sandpaper 1), $1.26 \pm 0.11N$ for the medium-friction surface (Sandpaper 2), and $0.61 \pm 0.07N$ for the low-friction surface (Glass). These results were consistent across participants ($mean \pm SD; n = 8$), indicating a clear and predictable relationship between the frictional properties of the surface and the force required to lift the object.

A one-way repeated-measures ANOVA was performed to statistically compare these peak forces across the three different friction conditions. The analysis revealed a significant difference between the frictional conditions ($F(2,21) = 92.46, p < .00001$), demonstrating that the type of surface material directly influenced the force dynamics during lifting. To further examine these differences, post hoc analyses using Bonferroni-corrected multiple comparisons were conducted. These tests confirmed that the peak tangential forces varied significantly between all three conditions, with a notable difference between high and medium friction surfaces ($High vs Medium, p = .00924$), and even more pronounced differences when comparing medium vs low ($p = .00000$) and high vs low ($p = .00000$) friction surfaces. These findings clearly demonstrate that the level of friction between the fingertip of the robotic sixth finger and the surface plays a critical role in determining the magnitude of tangential forces required to grip and lift the object. Higher friction surfaces, such as Sandpaper-1, required more substantial tangential forces to achieve successful lifting, as the increased resistance provided by the rough surface demanded greater force to overcome static friction. In contrast, the lower friction surface, like glass, required considerably less tangential force to initiate movement, as the smooth surface offered minimal resistance.

\begin{figure*}[th]       
    \includegraphics[width= 1\textwidth]{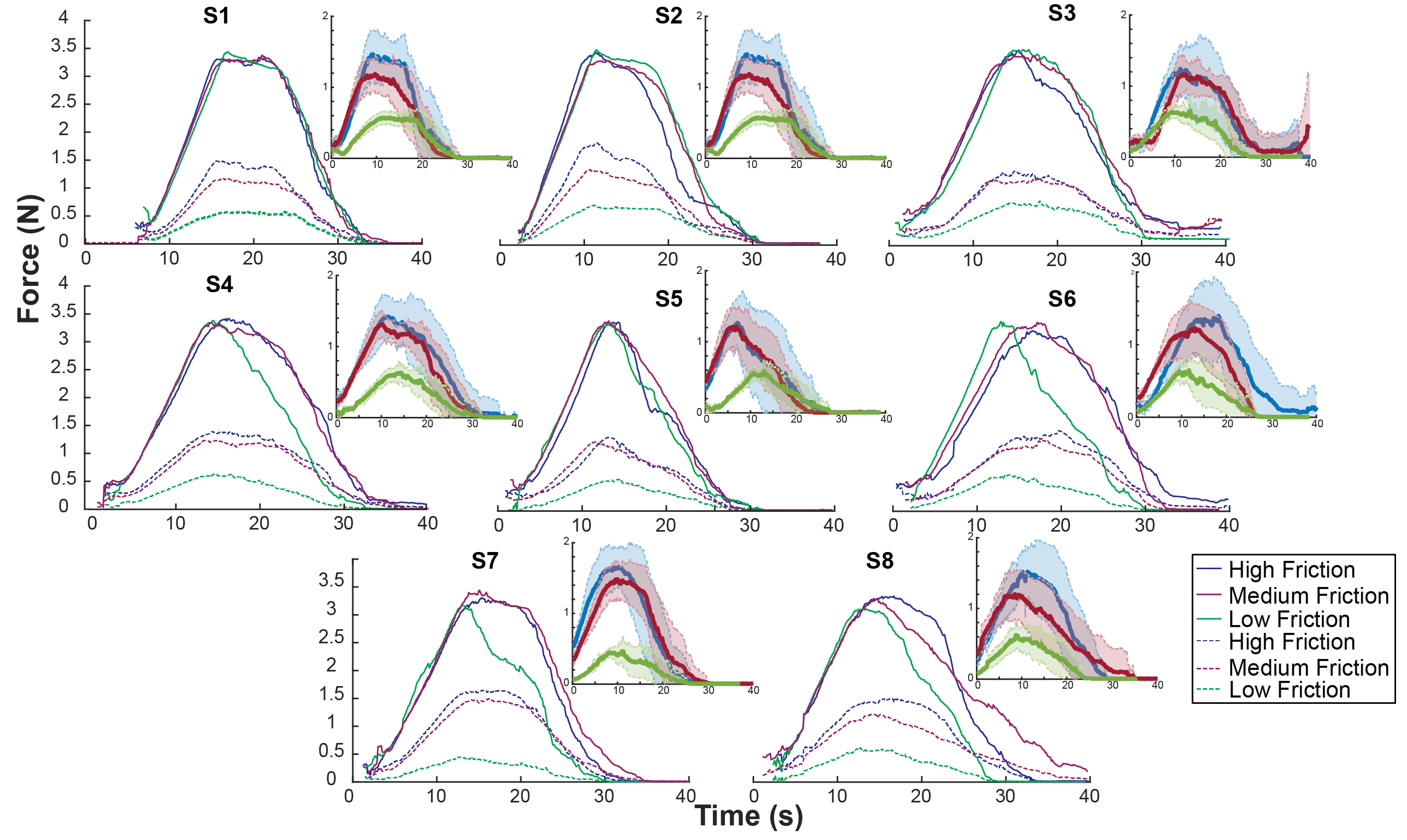}
    \centering
    \caption{Normal force and tangential force traces as a function of time: Each panel represents the average of the $20$ individual trials. Force traces are superimposed, solid lines indicate the normal force and the dotted line indicates the tangential force. The blue lines represent the forces during the high friction condition, the red lines represent the forces during the medium friction condition and the green lines represent the forces during the low friction condition. The inset in each graph represents only the mean tangential forces and the standard deviation synchronized at $0.5N$ for all the $20$ individual trials for all frictional conditions. The solid line indicates the mean and the shaded region indicates the standard deviation.}
\end{figure*}

\begin{figure}[!t]
\centerline{\includegraphics[width=\columnwidth]{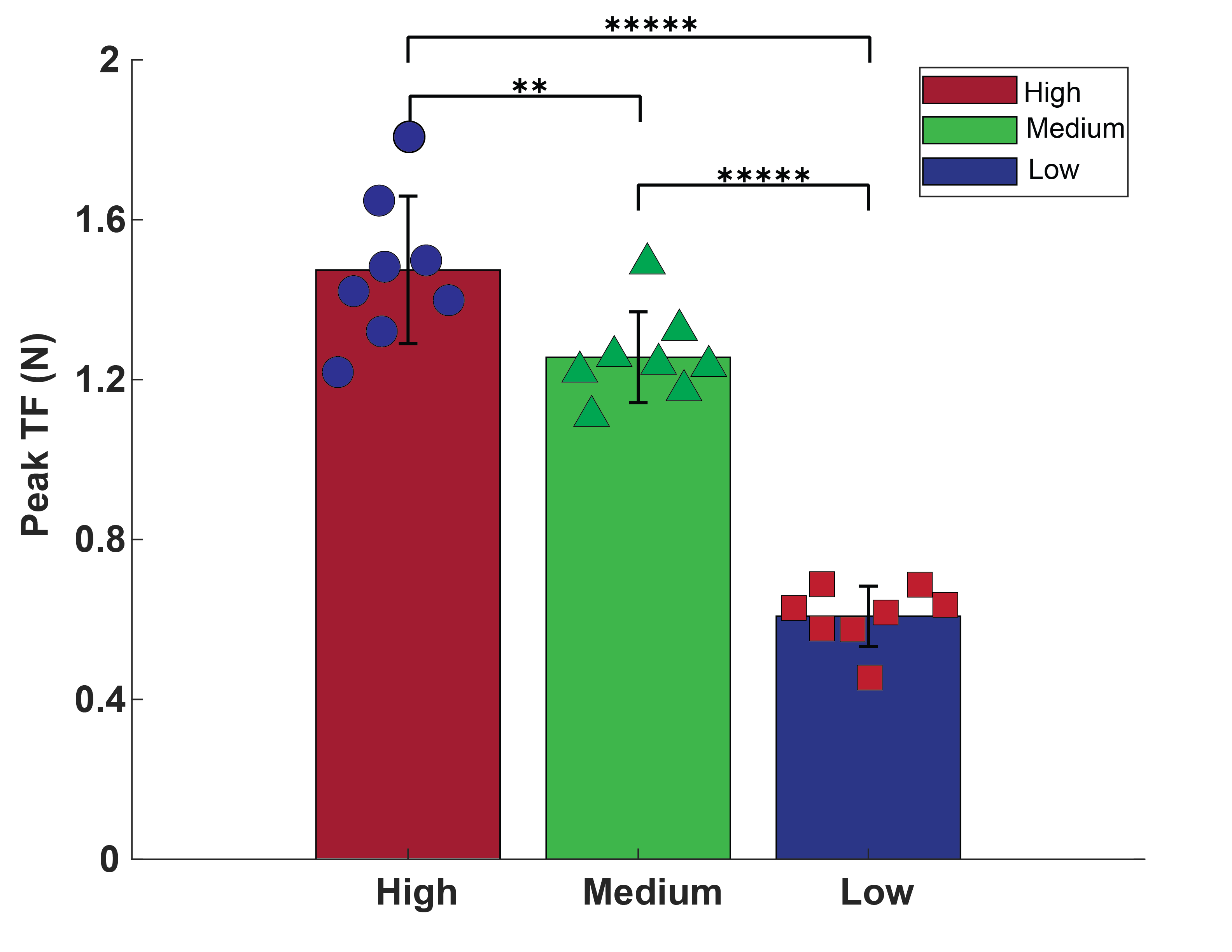}}
\caption{ Peak TF (tangential forces): Barplots displaying the mean and standard deviation of peak tangential forces across three frictional conditions. $*****P < = 0.000001$ (Bonferroni corrected). Blue circles show the average peak TF for each participant for High friction conditions, green triangles show the average peak TF for each participant for Medium friction conditions, and red squares show the average peak TF for each participant for low friction conditions}
\end{figure}

\subsection{Slip Detection via Second Derivative of Tangential Force Exceeding Threshold}

We analyzed the trials using the first condition check, where slip feedback was triggered by monitoring the second derivative of the tangential force, which corresponds to the acceleration of the force applied between the robotic sixth finger and the object. In this condition, slip feedback was initiated once the second derivative exceeded a predefined threshold, signaling a slip event. The slipperier the surface, the quicker the second derivative surpassed the threshold, leading to an earlier detection of slip. This resulted in lower frequency vibrations being delivered to the biceps, providing more rapid feedback to the user about the onset of the slip. Conversely, when the threshold was reached more gradually, as observed with higher friction surfaces, the frequency of the vibration motor increased, delivering stronger and more pronounced haptic feedback to the user’s biceps. This feedback system allowed the robotic sixth finger to effectively communicate slip events to the user, giving them an intuitive sense of how securely the object was being held, depending on the surface’s frictional properties. The dynamics of this process are illustrated in Fig. 7, where the slip ratio and the second derivative of the tangential force are plotted as functions of time. The left axis represents the slip ratio, which quantifies the extent of slip relative to the force applied, while the right axis shows the second derivative of the tangential force, representing the rate of change in the applied force. The graphs provide a clear visual representation of how slip develops over time under different friction conditions.

We plotted the average values of the slip ratio and the second derivative of the tangential force across the trials where slip feedback was triggered under the first condition check. Fig. 7(A) represents this phenomenon for the high-friction condition (Sandpaper-1), where the slip ratio increases more gradually over time, reflecting the higher resistance provided by the rough surface. This gradual increase results in a delay before the slip threshold is reached, leading to higher-frequency, stronger haptic feedback as the slip progresses more slowly. In contrast, Fig. 7(C) represents the low-friction condition (Glass), where the slip ratio increases rapidly once slip is initiated, leading to a more abrupt and pronounced slip event. In this case, once the slip begins, it continues until the object is fully dropped, as the smooth surface offers minimal resistance. This rapid onset of slip is accompanied by a lower frequency of vibrations, signaling a quicker detection of the slip event but with less intense feedback. Fig 7(B) represents the medium-friction condition (Sandpaper-2).

This comparison between high-friction and low-friction conditions underscores the importance of accurately detecting and responding to slip events, particularly when gripping objects with low-friction surfaces. In these cases, slip occurs more frequently and persists longer, as the robotic sixth finger struggles to maintain a stable grip. As a result, we opted to monitor latency, which is scaled by the coefficient of friction, as a more reliable indicator for triggering slip feedback. This approach ensures that slip events are detected and signaled promptly, especially in situations where high slip ratios persist, such as when manipulating objects with slippery surfaces. By incorporating this real-time slip feedback mechanism, the robotic sixth finger will enhance the user’s ability to adjust their grip dynamically and maintain control over objects across different surface types, effectively preventing drops and improving overall performance in gripping tasks.

\begin{figure}[t]
\centerline{\includegraphics[width=\columnwidth]{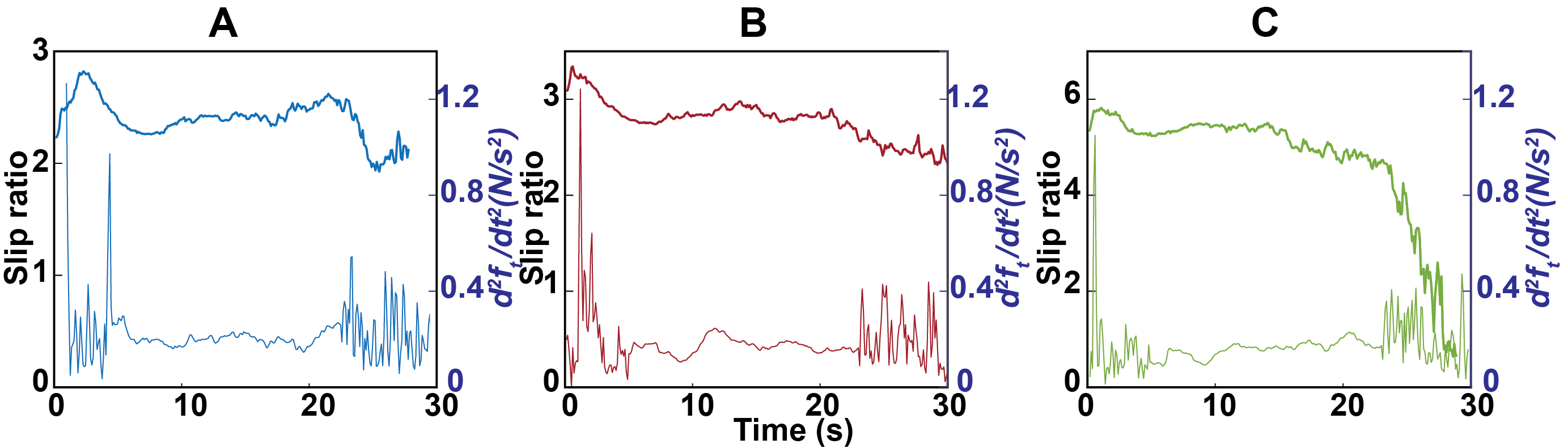}}
\caption{ Slip initiated at double-derivative threshold: The left axis of the graph has the slip ratio ({$f_n/f_t$}). The right axis is the double-derivative of the tangential force ({$f_t$}) which represents the acceleration. A, represented the slip ratio and double derivative for the high friction. B, represented the slip ratio and double derivative for the medium friction. C, represented the slip ratio and double derivative for the low friction.}
\end{figure}

\subsection{Slip Detection Based on Slip Ratio Difference Threshold}

We further analyzed the trials using the second condition check, where slip feedback was triggered based on the slip ratio difference threshold. This condition provided slip feedback when the difference between the peak slip ratio, which occurred during the grasping phase, and the instantaneous slip ratio exceeded the threshold. The feedback system responded to this difference by adjusting the intensity of the vibration delivered to the user, thereby indicating the degree of slip. Fig. 8 provides a detailed representation of the mean and standard deviation of the slip ratio across three frictional conditions high friction, medium friction, and low friction during trials where slip was detected using the slip ratio difference. At the start of each trial, after the sixth finger established a stable grasp on the object, the peak slip ratio was measured under varying frictional conditions. This peak represented the maximum slip ratio that occurred while the robotic finger maintained a secure hold on the object. The slip ratio was continuously monitored throughout the trial, tracking how the grip evolved over time.

Once the experimenter initiated the release phase of the sixth finger mimicking the scenario where the object was being dropped the difference between the peak slip ratio and the instantaneous slip ratio was calculated in real-time. This difference was critical in determining when and how the feedback system should respond. As the deviation between the peak and instantaneous slip ratios surpassed the preset threshold, the system activated, adjusting the vibration intensity delivered to the user’s biceps. The intensity of the feedback was inversely proportional to the magnitude of the peak slip ratio measured, with the larger peak slip ratio resulting in stronger intensity vibrations. This adaptive feedback mechanism ensured that the user was continuously aware of any subtle changes in the grasp stability, especially as the slip ratio fluctuated during the lifting and releasing phases. For high-friction surfaces, the peak slip ratio was relatively small, resulting in high-intensity vibrations. On the other hand, for low-friction surfaces, such as glass, the peak slip ratio was larger, leading to low-intensity feedback vibration. This highlighted the system’s ability to detect the more frequent and pronounced slips characteristic of low-friction conditions, where maintaining grip stability is more challenging.

\begin{figure}[!t]
\centerline{\includegraphics[width=\columnwidth]{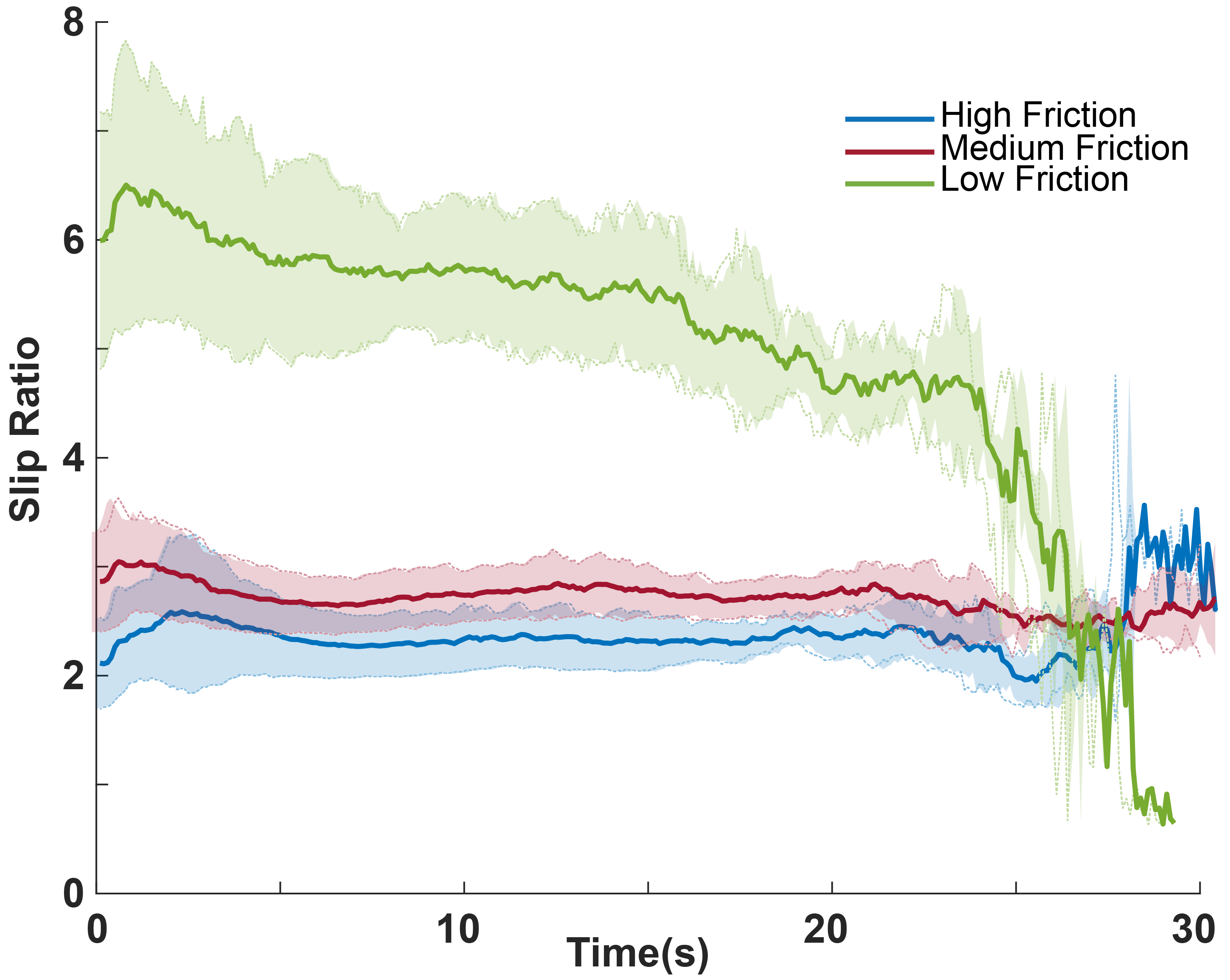}}
\caption{ Slip ratio (SR): The solid lines indicate the average slip ratio of all the trials. The shaded region indicates the standard deviation. The blue line indicates the slip ratio for the high friction condition, the red line indicates the slip ratio for the medium friction condition and the green line indicates the slip ratio for the low friction condition.}
\label{fig1}
\end{figure}

\section{Discussion}
\label{sec:guidelines}

The utilization of extra-robotic limbs to enhance human motor abilities is extremely useful, but it also poses significant challenges to be completely integrated. One of the biggest hurdles is the absence of the sensory feedback we naturally rely on from our limbs. Without sensations like touch, pressure, or the awareness of our body's position (proprioception), it is challenging for the users to control these robotic limbs smoothly and naturally. This gap in sensory feedback makes it hard to truly incorporate these limbs into a person’s everyday movements, limiting their potential. To transform extra-robotic limbs from an assistive device for limb rehabilitation to be a fully functional limb, we need to create advanced sensory systems that can deliver real-time feedback in a way that feels natural, almost like a body part. If we can achieve that, we will be able to unlock the full potential of these technologies, allowing them to significantly enhance human capabilities. In this study, we set out to bridge that gap by developing a novel technique aimed at providing users with frictional feedback when they try to lift objects and encounter slippage with these robotic limbs. Grasping and manipulating objects with varying surface textures and friction requires this kind of information for a secure grip. In humans, tactile feedback from our fingers provides crucial frictional information, which we sense either through exploratory finger movements across a surface or when manipulating an object and feeling it start to slip \cite{https://doi.org/10.1113/JP286027, doi:10.1152/jn.00504.2020}. This feedback about friction is extremely important for maintaining a stable grip in order to avoid dropping off the objects and for surface exploration. Without it, even the most advanced robotic limb will struggle to mimic the natural dexterity of the human hand. 
The robotic sixth finger was originally designed to assist stroke patients by compensating for lost hand function, rather than serving as a comprehensive tool for hand rehabilitation or a fully functional hand replacement. Its primary goal was to provide support for basic tasks, rather than restoring the full range of motion and sensory feedback that a natural hand offers. As a result, earlier studies did not attempt to equip the device with sensory feedback or automatic force control. However, in this study, our goal is to integrate the capability to provide sensory feedback as a first step. The idea is that this feedback will eventually be used either for manual adjustments by the user or automatic force control by the device itself. Integrating friction feedback into these extra-robotic limbs, we believe we can make a significant step forward in enabling smoother, more intuitive control, allowing users to interact with their environment in a much more natural way. The long-term vision is that, with sensory feedback systems like this, robotic limbs will become more than just functional tools, they will start to feel like real extensions of the user’s body.

Previous studies have demonstrated that vibrotactile feedback (VTF) is well-suited for providing supplemental sensory information without taxing users' auditory or visual attention, and without interfering with other essential functions. This makes VTF an ideal choice for applications where additional sensory input is needed, such as in assistive devices or robotic prosthetics. Furthermore, VTF has practical advantages it is non-invasive, meaning it does not require complex surgical or physical modifications, and it’s also relatively cost-effective to implement, making it a viable option for wider use in rehabilitation technology \cite{10.3389/fresc.2022.895036, wang2019haptics, 7558904, 10912496}. 
Given these benefits, we chose to use VTF to convey critical information about friction and to cue the onset of slippage for users attempting to lift objects with a robotic sixth finger. This was particularly important for our target application, as the device was designed to assist stroke rehabilitation patients. To replicate this scenario in our experiments, we simulated a situation where the dorsal region of a paretic (weakened) hand acts as a support. At the same time, the robotic sixth finger is instrumented with a force sensor at its fingertip which gets in contact with the object. Haptic feedback was provided on the same limb, specifically on the biceps of the same limb, using a single vibration motor, ensuring that only one limb was engaged in the task. 

To detect slippage, we developed an algorithm based on two key parameters: the second derivative of the tangential force (which we refer to as slippage acceleration) and the slip ratio difference. This allows the algorithm to adjust the intensity of the vibration motor based on friction levels, providing the user with real-time, precise feedback. The second derivative of the slip ratio is critical for capturing how quickly slippage occurs. For low-friction objects, slippage tends to accelerate more rapidly, while high-friction objects experience slower slippage due to increased resistance. By monitoring these changes, the system can identify different friction levels, enabling users to adjust their grip more effectively. The second component of the algorithm, the slip ratio difference, acts as an additional safeguard. In some cases, the first parameter alone might not detect slippage, especially when the hand is partially supporting the object. Here, the object might still slip at the fingertip, but the rate of slippage may not be significant enough to trigger a response based on slippage acceleration alone. However, the frictional effects still need to be addressed, and this is where the slip ratio difference comes into play. It provides a backup mechanism, ensuring that feedback is delivered even if the initial condition doesn’t trigger the response. To account for the diverse behaviors of slippage and friction across different objects and environments, we further refined the algorithm to avoid oversimplifications. The updated version continuously monitors the slip ratio in real time, allowing it to detect even subtle variations in friction. This dynamic feedback system instantly alerts the user when grip adjustments are necessary, ensuring a more natural and intuitive experience. This advanced approach not only enhances rehabilitation efforts for stroke patients but also paves the way for broader applications in prosthetics and assistive devices. By offering a non-invasive, real-time method for conveying frictional information, we aim to make robotic limbs feel like true extensions of the user's body. This improvement allows users to perform tasks with greater precision and confidence, making these devices more practical and effective in everyday life.

The ability to perceive friction through vibrotactile cues across three distinct friction conditions was impressive, with all participants achieving over 90\% accuracy across all 30 trials. We used a single-vibration motor to focus exclusively on delivering slip and friction information by varying the intensity of the vibrations. This streamlined approach effectively communicated the necessary sensory feedback, enabling subjects to differentiate between friction levels with high precision. The initial touch was signaled by a single vibration to inform the user that the object had made contact, mimicking the function of fast-adapting tactile afferents (FA-I). Similar to how FA-I afferents respond rapidly to the onset of touch (contact timing) but then quickly adapt and become inhibited later \cite{articleJS1987}, the single vibration serves as an immediate alert without overwhelming the user with continuous feedback. This approach allows the user to recognize the contact moment while maintaining focus on subsequent actions. The confusion matrix (Table 2) shows that most confusions occurred when distinguishing the intermediate friction level, while the other friction levels were easily differentiated with high accuracy. Localization of the vibration motor was chosen carefully to maximize the perceptual accuracy \cite{8013825}. These results highlight the effectiveness of the stimulation site and the vibrotactile modality in conveying sensory information. Augmenting the robotic sixth finger with sensory feedback represents a significant step toward transforming it from a basic assistive device \cite{1501098} into a more advanced, intuitive tool. This enhancement improves the user’s ability to interact with objects naturally, bringing the device closer to functioning as an integrated extension of the body. 

In order to validate the study, we intentionally kept the grip force constant throughout the experiments. Our primary aim was to assess the effectiveness of using vibrotactile feedback to convey sensory information, so we controlled this variable. After conducting several pilot trials, we determined that a grip force of 3.5N was sufficient to lift the glass across all three friction conditions. This value was selected as it provided consistent performance across the different friction levels. Tangential force analysis showed that the patterns of grip force regulation and slip closely resembled those seen in precision grip scenarios with small objects \cite{johansson1987signals}. Johansson \textit{et al.} demonstrated the load force levels at which slippage occurs by plotting the acceleration record against load force \cite{johansson1984roles}. In our study, we used the second derivative of tangential force to detect slippage, which offered a highly accurate measure of when slippage occurred. A shorter latency in surpassing the threshold indicated a more slippery surface, while a longer latency corresponded to a higher friction surface. This method enabled us to assess frictional properties based on the timing of slippage detection, giving us more insight into the interaction between grip force and surface texture. The ability to sense friction with the robotic sixth finger goes beyond simply controlling grip force. It also provides valuable sensory information about surface properties and textures, significantly enhancing the sensory capabilities of these additional fingers. This enriched feedback can lead to more refined and effective interactions with various surfaces, improving the overall functionality of the device.

In this paper, we aimed to integrate three key parameters of sensory information into the robotic sixth finger during interactions with surfaces and while gripping and lifting objects: contact timing, frictional information, and incipient slip. Drawing inspiration from the tactile afferents embedded in human skin, we modeled our approach after the response patterns of these afferents to various contact events \cite{johansson2009coding}. One limitation of our study is that it does not address the precise manipulation of objects, which was not the primary focus of the prototype we developed. Instead, our main objective was to enable the device to sense slip and friction and to effectively relay this sensory information to the user. By emphasizing these aspects, we lay the groundwork for future developments in the system. Our primary goal is to enhance the functionality of the robotic sixth finger by programming it to utilize slip information for automatic grip force adjustments and more refined manipulation of objects. Additionally, we aim to enhance the capabilities of the robotic sixth finger by enabling it to sense the surface textures of objects and relay this information to the user, much like a natural finger would. This feature would provide users with crucial insights into the tactile properties of various surfaces, such as smoothness, roughness, or stickiness. By incorporating texture sensing, the robotic finger would not only improve grip stability but also enhance the user's overall tactile experience. This sensory feedback could help users make more informed decisions when interacting with different materials, allowing for greater precision in tasks that require fine motor skills. Ultimately, this development would further bridge the gap between robotic and biological systems, enabling a more seamless integration of the robotic finger into everyday activities. This advancement would not only improve user experience but also enable the device to operate more autonomously, allowing users to interact with objects in a more natural and effective manner. Ultimately, we envision a future where this technology provides comprehensive sensory feedback, making robotic limbs feel like an integral part of the user's body.

\section{Conclusion}
The work presented in this paper represents an important initial step toward enhancing the sensory capabilities of extra fingers, transitioning them from mere assistive devices to fully functional extensions of the hand. By utilizing vibrotactile feedback to encode information about friction and slip, we provide users with essential sensory input while lifting lightweight objects. This is the first study, offering a novel technique to advance sensory feedback technologies for rehabilitation and the enhancement of motor capabilities. Our primary goal is to further improve the functionality of the robotic sixth finger, ultimately enabling it to explore the surface properties of various objects. By doing so, we aim to create a more intuitive and effective interface that closely mimics the natural capabilities of human fingers, thereby enhancing the user's experience and interaction with their environment.

We anticipate that this development will enhance performance, particularly in the manipulation of objects, leading to greater independence in activities of daily living (ADLs). By providing users with improved sensory feedback, we aim to reduce reliance on erroneous compensatory motor strategies that often arise when tackling everyday tasks. This focus on refined control will empower users to interact with their environment more effectively and confidently \cite{ROBYBRAMI2003217}. We are actively working on further refining the slip detection algorithm to enable complete regulation of grip force, either manually by the users or automatically. In humans, most adjustments in grip force occur instinctively, and we aim to replicate this natural response in our device. By enhancing the algorithm, we hope to provide users with an experience that allows for intuitive control and improved interaction with objects, ultimately making the robotic finger function more like a biological one. \cite{doi:10.1152/jn.1998.80.4.1989}. We are also exploring additional solutions for controlling these fingers by leveraging electroencephalographic (EEG) activity to drive voluntary actions. This innovative approach could enable users to control the robotic fingers through their brain activity, creating a more intuitive interface. Additionally, we are considering the integration of functional electrical stimulation (FES) to activate nerve fibers, providing real-time sensory feedback. By combining EEG-driven control with FES, we aim to enhance the user experience, allowing for more natural and responsive interactions with objects. We are also exploring the potential of using our robotic extra finger for patients with various neurological conditions that impact hand grasping. This includes diseases such as Multiple Sclerosis, Amyotrophic Lateral Sclerosis, and paresis resulting from cervical spinal cord lesions.

\section*{Acknowledgment}
This work was supported by Khalifa University Center for Autonomous and Robotic Systems under Award RC1-2018-KUCARS, MBZIRC-8434000194, KU-BIT-Joint-Lab-8434000534, and the Advanced Research and Innovation Center (ARIC), which is jointly funded by Mubadala UAE Clusters and Khalifa University of Science
and Technology.

\section {Conflict of interest}
The authors declare that they have no competing financial interests.

\bibliographystyle{elsarticle-num} 
\bibliography{Final_Draft}





\end{document}